\documentclass[runningheads]{llncs}
\usepackage{float}
\usepackage{graphicx}
\usepackage{amsmath}
\usepackage{xcolor}
\usepackage{array}
\usepackage{mwe}
\usepackage{subfig}
\usepackage{xr}
\usepackage[misc,geometry]{ifsym}
\makeatletter
\newcommand*{\addFileDependency}[1]{
  \typeout{(#1)}
  \@addtofilelist{#1}
  \IfFileExists{#1}{}{\typeout{No file #1.}}
}
\makeatother

\def\perturbator {M}
\def\encoder {f}
\def\classif {p}
\def\operator {T_{\Lambda_M}}
\def\projection {g}
\def \information {I_{NCE}}
\DeclareMathOperator{\erfinv}{erfinv}

\begin{document}

\title{Optimizing transformations for contrastive learning in a differentiable framework}
\titlerunning{Differentiable transformations in contrastive learning}

\author{Camille Ruppli \inst{1,3}
\and
Pietro Gori\inst{1} \and
Roberto Ardon\inst{3} \and
Isabelle Bloch \inst{2,3}}

\authorrunning{C. Ruppli et al.}

\institute{LTCI, Télécom Paris, Institut Polytechnique de Paris, Paris, France \and
Sorbonne Universit\'e, CNRS, LIP6, Paris, France \and
Incepto Medical, Paris, France 
}

\maketitle             

\begin{abstract}
Current contrastive learning methods use random transformations sampled from a large list of transformations, with fixed hyper-parameters, to learn invariance from an unannotated database. 
Following previous works that introduce a small amount of supervision, 
we propose a framework to find optimal transformations for contrastive learning using a \textit{differentiable} transformation network.
Our method increases performances at low annotated data regime both in supervision accuracy and in convergence speed. In contrast to previous work, no generative model is needed for transformation optimization. Transformed images keep relevant information to solve the supervised task, here classification.
Experiments were performed on 34000 2D slices of brain Magnetic Resonance Images and 11200 chest X-ray images. On both datasets, with 10\% of labeled data, our model achieves better performances than a fully supervised model with 100\% labels.

\end{abstract}

\section{Introduction}

When working with medical images, data are increasingly available but annotations are fewer and costly to obtain. Self-supervised methods have been developed to take full advantage of the non-annotated data and increase performances in supervised tasks at low annotated data regime.
As part of self-supervised methods, contrastive learning methods \cite{NEURIPS2020_949686ec,pmlr-v119-chen20j,Patrick_2021_ICCV,perakis_contrastive_2021} train an encoder on non-annotated data to learn invariance between transformed versions of images. 
Contrastive learning methods are also used with medical images.
For instance, the authors of~\cite{NEURIPS2020_949686ec} learn local and global features invariance while those of~\cite{Dufumier2021ContrastiveLW} introduce a kernel to take metadata into account in contrastive pretraining.

In most works, the transformations used to learn invariance are randomly sampled from a given list.
While many works study the impact of removing some transformations on supervised task performance~\cite{pmlr-v119-chen20j,perakis_contrastive_2021}, not much investigation has been done on optimizing the transformations and their hyper-parameters. Some authors \cite{Patrick_2021_ICCV,xiao2021what} focus on the role of transformations but without explicit transformations optimization. The work of \cite{Patrick_2021_ICCV} proposes a formal analysis of transformations composition to select admissible transformations while \cite{xiao2021what} explores the latent spaces of specific transformations.
The authors of \cite{Yang2021DistributionET} introduce a generative network to learn transformations distribution present in the data to use complementary transformation in self-supervised tasks. Unlike our work (see Section~\ref{sec:method}) they need a pretraining step before the contrastive one to learn transformations distribution.

Within supervised training (not self-supervision), some works have proposed to optimize data augmentation. In~\cite{cubuk_autoaugment_2019}, a pre-training step using reinforcement learning is required. The work of~\cite{zhao_differentiable_2020} shows that data augmentation should be applied on both discriminator and generator optimization steps but no optimization is performed on augmentation choice.
The authors of~\cite{li_differentiable_2020,liu_direct_2021} learn a vector containing augmentations probability. They also present a transformations optimization strategy. Unlike our approach (see Section~\ref{sec:method}), transformation parameters are discretized. Optimization is performed on the probability of choosing a family of transformations and a set of parameters.

While supervision is also introduced in contrastive learning in~\cite{NEURIPS2020_d89a66c7,Zhao_2021_ICCV}, few authors used it in order to influence the choice of transformations. Among them, the authors of~\cite{tian_what_2020} introduce a transformation generator (a flow-based model based on~\cite{kingma_glow_2018}) to generate transformed images in new color spaces minimizing mutual information while keeping enough information for the supervised task. 
As transformations only impact color spaces, their application to gray scale images, in particular medical images, is very limited. Furthermore, consistently synthesizing anatomically relevant images with generative models can be difficult \cite{hallucinate}. To the best of our knowledge, \cite{tian_what_2020} is the only existing method optimizing a transformation generator for contrastive learning.

As in~\cite{tian_what_2020}, the present work uses a small amount of supervision (10\%) for transformation optimization. We introduce a differentiable framework on transformations that needs no pre-training, and, unlike~\cite{tian_what_2020}, is applicable to both color and gray scale images.
Our contributions are the following:
\begin{itemize}
    \item  We propose a semi-supervised differentiable framework to optimize the transformations of contrastive learning.
    \item We demonstrate that our method finds relevant transformations for the downstream task, which are easy to interpret.
    \item We show that our framework has better performances than fully supervised training at low data regime and contrastive learning~\cite{pmlr-v119-chen20j} without supervision.
\end{itemize}

\section{Transformation network}\label{sec:method}

Contrastive learning methods train an encoder to bring close together latent representations of positive pairs of images while pushing away representations of negative pairs of images. As in simCLR~\cite{pmlr-v119-chen20j}, positive pairs are two transformations of the same image while negative pairs are transformed versions of different images.

Transformations used in most methods are chosen at random from a fixed given list. However, as shown in \cite{tian_what_2020}, using positive (transformed) images, that are very similar to each other (i.e., high mutual information), might entail a sub-optimal solution since it would not bring additional information to the encoder. By using a small amount of supervision, transformations can be optimized in order to contain relevant information for the targeted supervised task

\begin{figure}
\centering
\includegraphics[scale=0.75]{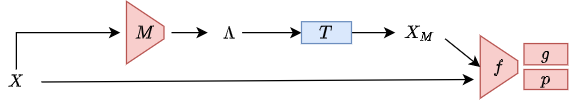}
\caption{Proposed architecture (red color indicates a trainable element, blue color indicates a non-trainable element).} \label{simclr_M_enc_architecture}
\end{figure}

In this work, we focus on classification tasks.
We introduce a transformation network ($\perturbator$) that minimizes the mutual information between images of a positive pair without compromising the supervised task performance.
For each image of the training set, $\perturbator$, implemented as a neural network, outputs a set of parameters ($\Lambda$) defining the transformations to apply ($\operator$). As in~\cite{pmlr-v119-chen20j,xiao2021what}, the latent space of the encoder ($\encoder$) is optimized using a projection head ($\projection$) into a lower dimension space  where a contrastive loss function ($\information$)  is minimized. Supervision is added on the latent space using a linear classifier ($\classif$) that minimizes a classification loss function ($\cal{L}$). 
Fig.~\ref{simclr_M_enc_architecture} shows a schematic view of the architecture used ($X$ denotes an image from the training set and $X_M$ its transformed version). 

\subsection{Optimizing transformations}

We consider a finite set of intensity and geometric transformations acting on images.
Each transformation is parameterized by a vector of parameters (for example, the parameter vector of a rotation around a fixed point only contains its angle).
The transformation function ($\operator$) is the composition of transformations applied in a fixed order.
The transformation network ($\perturbator$) outputs the transformation function parameters. We propose to train $\perturbator$ to find the optimal transformations for the semi-supervised contrastive problem. The network $\perturbator$ maps an image to the space of parameter vectors, normalized to $[0,1]$. The order of the transformations in the composition is not optimized, but the impact of this order has been studied and results are shown in Section~\ref{results}.

Let $\lambda_k$ be the vector of parameters for a given transformation, then the transformation function, noted as  $\operator$, is parameterized by $\Lambda = [\mathbf{\lambda}_1, \cdots, \mathbf{\lambda}_K]$ (where $K$ is the number of transformations considered).

The optimal transformations for the semi-supervised contrastive problem is then obtained via $\perturbator$, which is thus responsible for finding the optimal $\Lambda^*_M$. In contrast with~\cite{pmlr-v119-chen20j}, we only transform one version of the image batch. Our experiments show  better results in this setting. The optimization goes as follows.

\noindent \textbf{Transformation network optimization steps:}
(i)~$\perturbator$ generates a batch of $\Lambda_M$ vectors defining a transformation $T_{\Lambda_M}$. For every image $X$ in a batch, a transformed version is generated: $X_M = T_{\Lambda_M}(X) $. 
(ii)~The transformed and untransformed data batches are passed through the encoder $\encoder$, the projection head~$\projection$ and the linear classifier $\classif$.
(iii)~The contrastive loss $-\information$ (see below, Eq.~\ref{eq:INCE}) gradient is computed to update the weights of the network $\perturbator$ aiming to minimize mutual information and classification loss function.

\noindent \textbf{Encoder optimization steps:} 
(i)~From the previous optimization steps of~$\perturbator$, one transformed version of the data is generated. Latent projections of the transformed and untransformed data are generated using encoder $\encoder$ and projection head~$\projection$.
(ii)~The contrastive loss gradient is computed and parameters of $\encoder$, $\projection$ and $\classif$ are updated. This brings closer positive pairs and further away negative ones, and ensures that transformed images are properly classified.

Formally, these steps aim to solve the following coupled optimization problem where contrastive and classification loss functions are taken into account:

\begin{equation}
\label{eq:minmax}
\left\{
\begin{array}{cl}
     \min_\perturbator & \alpha_0\information \Big(\projection \circ \encoder(X_M), \projection \circ \encoder(X) \Big) + \alpha_1{\cal{L}}\Big(\classif \circ \encoder(X_M),y \Big)  \\
    \min_{\encoder, \classif, \projection}& - \alpha_2\information \Big(\projection \circ \encoder(X_M), \projection \circ \encoder(X) \Big) + \alpha_3{\cal{L}}\Big(\classif \circ \encoder(X_M\big),y \Big)  \\
        & + \alpha_4{\cal{L}}\Big(\classif \circ \encoder(X), y \Big)
\end{array}
\right.
\end{equation}
where $\alpha_i$ are weights balancing each loss term and $y$ are the classification labels when available.
The terme $\information$ is the contrastive loss function as in~\cite{pmlr-v119-chen20j}: 
\begin{equation}
\label{eq:INCE}
\information({X_M}_i,X_i) = - \sum_{i}\log\left(\frac{e^{sim(\projection(\encoder({X_M}_i)), \projection(\encoder(X_i)))}}{\sum_{j, j\neq i} e^{sim(\projection(\encoder({X_M}_i)), \projection(\encoder(X_j)))}}\right) \
\end{equation}
where the index $i$ defines positive pairs, $j$ negative ones, and  $sim$ is a similarity measure defined as $sim(x,x') = \frac{x^Tx'}{\tau}$ where $\tau$ is a fixed scalar, here equal to 1. Finally, $\cal{L}$ is the binary cross entropy loss function for the supervised constraint.

\subsection{Differentiable formulation of the transformations}

A fundamental difference of the proposed transformation optimization, compared to \cite{li_differentiable_2020,liu_direct_2021,tian_what_2020}, is the use of explicit transformations differentiation.
During training, gradient computations of Eq.~\ref{eq:minmax} involve the derivative of $\operator$ with respect to the weights ($w$) of $\perturbator$:
$d_w(T_{\Lambda_M}) = dT_{\Lambda_M} \circ d_w M$. This requires the explicit computation of the derivatives of $T$ with respect to its parameters $\Lambda$ and the differential calculus for each transformation composing $T$. 
Thus, we introduce specific formulations and normalized parameterization for the transformations used in our experiments.

We use the following transformations: crop ($Crop$), Gaussian blur ($G$), additive Gaussian noise ($N$), rotation ($R$) around the center of the image, horizontal ($Flip_0$) and vertical ($Flip_1$) flips.
Table~\ref{transformations} lists the expressions of these transformations.  
The final transformation function is defined as:
\begin{equation}
  T_\Lambda = (R \circ Flip_1 \circ Flip_0 \circ Crop \circ N \circ G)(X,\Lambda)  
\end{equation}
and $T_\Lambda$ thus depends on 7 parameters (the crop has 2 parameters) which are generated by $\perturbator$.
\begin{table}
\caption{\small Differentiable expressions of the transformations used, parameterized by $\lambda \in [0,1]$, where $S$ is the sigmoid function, $s$ the size of our images, $\erfinv$ the inverse of the error function $2\pi^{-\frac{1}{2}}\int_x^\infty e^{-u^2}du$, $\cal{U}$ the uniform distribution and $x$ is a point of the image grid. We fix the maximum Gaussian blur standard deviation to $\sigma_{max} = 2.0$ and the maximum additive noise standard deviation to $\Tilde{\sigma}_{max} = 0.1$.}
\label{transformations}
\begin{center}
\begin{tabular}{|l|l|}
\hline
Flip around axis $e$ & $Flip(X, \lambda, e)(x) = (1 - \lambda)X(x) + \lambda X(x - 2 \langle x, e \rangle e)$  \\
\hline
Crop centered at $c_\lambda = [\lambda_1 s, \lambda_2 s]$ & $Crop(X, \lambda)(x) = X(x) \times S(\frac{s}{8} - ||x - c_\lambda||_{\infty})$ \\
\hline
Gaussian blur with kernel $g_{\lambda\sigma_{max}}$ & $G(X,\lambda) = g_{\lambda\sigma_{max}} * X$ \\
\hline
Rotation      &         $R(X, \lambda)(x) = 
X \left(
\begin{pmatrix}
 \cos(\lambda2\pi) & -\sin(\lambda2\pi)\\
 \sin(\lambda2\pi) & \cos(\lambda2\pi)
\end{pmatrix}x\right)$  \\
\hline
Additive Gaussian noise &        $N(X,\lambda) = X + \lambda\Tilde{\sigma}_{max} \times \sqrt{2}\erfinv({\cal{U}}[-1,1]) $ \\
\hline
\end{tabular}
\end{center}
\vspace{-5mm}
\end{table}

\subsection{Experimental settings}
\textbf{Dataset} Experiments were performed on BraTs MRI~\cite{6975210} and Chest X-ray~\cite{Tang2020AutomatedAC} datasets.
The Chest X-ray dataset is composed of 10000 images.
BraTs volumes were split along the axial axis to get 2D slices. Only slices with less than 80\% of black pixels were kept. This resulted in 34000 slices. For both datasets, we studied the supervised task of pathology presence classification (binary classification, present/not present). In medical imaging problems, it is common to have labels only for a small part of the dataset.
We thus choose 10\% of supervision in all of our experiments.
We randomly selected three hold-out test sets of 1000 slices for BraTs experiments. With the Chest dataset, we used the provided test set of 1300 images, from~\cite{Tang2020AutomatedAC}, evenly split in three to evaluate variability.

\noindent \textbf{Implementation details} For every experiment with the BraTs dataset, the encoder $\encoder$ is a fully convolutional network composed of four convolution blocks with two convolutional layers in each block. 
Following the architecture proposed in~\cite{Tang2020AutomatedAC}, the encoder $\encoder$ for experiments on the Chest dataset is a Densenet121.
The network $\perturbator$ is a fully convolutional network composed of two convolutional blocks with one convolutional layer.
The projection head~$\projection$ is a two-layer perceptron as in \cite{pmlr-v119-chen20j}. 
On BraTs dataset (resp. Chest dataset), we train with a batch size of 32 (resp. 16) for 100 epochs. In each experiment, the learning rate of $\encoder$ is set to $10^{-4}$. When optimizing $\perturbator$ with (resp. without) supervision, $\perturbator$ learning rate is set to $10^{-3}$ (resp. $10^{-4}$).
When using 10\% of labeled data for the supervision task, on relatively small databases ($10^5$ images), there is a risk of overfitting on the classification layer ($\classif$ in Eq.~\ref{eq:minmax}). Contrastive and supervision loss terms need to be carefully balanced while optimizing both the encoder and the transformation generator. To evaluate the impact of hyper-parameters, we carried out experiments with ($\alpha_0, \alpha_2) \in \{1, 0.1\}$ and $(\alpha_1, \alpha_3, \alpha_4) \in \{1, 10\}$. Linear evaluation results (see Section~\ref{linear_evaluation}) on BraTs dataset after convergence are summarized in Table~\ref{hyperparam_results}.
Results in Section~\ref{results} are shown with the best values found for each method.

\begin{table}
\caption{ \small 3-fold cross validation mean linear evaluation AUC after convergence with different $\alpha_i$ values (standard deviation in parenthesis).}
\label{hyperparam_results}
\begin{center}
\begin{tabular}{|l|l|l|}
\hline
& $\alpha_i$ values & AUC  \\
\hline
Optimizing M & $\alpha_{0,2} = 1, \alpha_{3,4} = 1, \alpha_1=10$ & \textbf{0.884} (0.042)  \\
& $\alpha_0 = 0.1, \alpha_{1,3,4} = 10, \alpha_2 = 0.1$ & 0.868 (0.030)\\
& $\alpha_0 = 0.1, \alpha_1 = 10, \alpha_2 = 1, \alpha_{3,4} = 1$ & \textbf{0.887} (0.013) \\
\hline
Random M & $\alpha_2 = 1, \alpha_{3,4} = 1$ & 0.874 (0.000) \\
& $\alpha_2 = 0.1, \alpha_{3,4} = 10$ & 0.820 (0.037) \\
 & $\alpha_2 = 1, \alpha_{3,4} = 10$ & \textbf{0.883} (0.003) \\
 \hline
base simCLR~\cite{pmlr-v119-chen20j} & & 0.730 (0.020) \\
\hline
\end{tabular}
\end{center}
\vspace{-5mm}
\end{table}

The fully supervised experiments described in Section~\ref{results} are optimized with the same encoder architecture and one dense layer followed by a sigmoid activation function for the classification task. For the fully supervised experiments we used a learning rate of $10^{-4}$.
\\
\textbf{Computing infrastructure} Optimizations were run on Tesla NVIDIA V100 cards.

\subsection{Linear evaluation}\label{linear_evaluation}
To evaluate the representation quality learned by the encoder, we follow the linear evaluation protocol used in the literature~\cite{pmlr-v119-chen20j,perakis_contrastive_2021,tian_what_2020}. The encoder is frozen with the weights learned with our framework. One linear layer is added, after removing the projection head ($\projection$), and trained with a test set of labeled data, not used in the previous training phase. This means that we first project the test samples in the latent space of the frozen model and then estimate the most discriminative linear model. The rationale here is that a good representation should make the classes of the test data linearly separable.

\section{Results and discussion}\label{results}

To assess the impact of each term in Eq.~\ref{eq:minmax} we performed optimization using the following strategies:\\
\textbf{Random} (without $\perturbator$, without supervision): each image is transformed with parameters generated by a uniform distribution: $\Lambda = {\cal{U}}\left([0, 1]^7\right)$, and $\alpha_{1,3,4} =  0$.\\
\textbf{Random with supervision} (without $\perturbator$, with supervision): we add the supervision constraint to the random strategy. We set $\alpha_2 = 1$ and $\alpha_{3,4} = 10$.\\
\textbf{Self-supervised} (with $\perturbator$, without supervision): while setting $\alpha_1$, $\alpha_3$ and $\alpha_4$ to 0, we optimize Eq.~\ref{eq:minmax}.\\
\textbf{Self-supervised with supervision constraint} (with $\perturbator$ and supervision): setting $\alpha_1 = 10$ and $\alpha_{0,2,3,4} = 1$, we optimize Eq.~\ref{eq:minmax}.  

 We split the data into pre-training and test sets. Data from the pre-training set are further split into training and validation sets for the perturbator/encoder optimization. 
 For optimizations with supervision constraint (self-supervised and random), all pre-training data are used for self-supervision and a small set of labeled data is used for the supervision constraint. For variability analysis, three optimizations were performed by changing the supervision set.
With the BraTs dataset, as slices come from 3D volumes, we split the data ensuring that all slices of the same patient were in the same set.

\begin{figure}
\includegraphics[width=\textwidth]{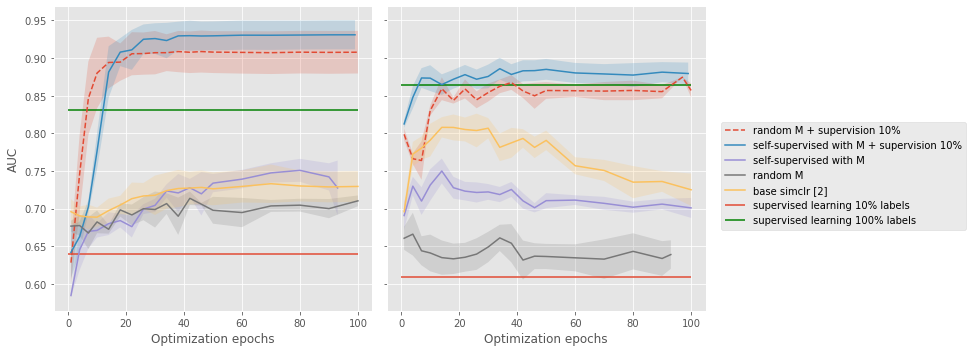}
\caption{Linear evaluation results comparing with other methods (left BraTs dataset with batch size 32, right Chest dataset with batch size 16).} \label{4methods_comparison}
\end{figure}

\noindent \textbf{Linear evaluation } was performed on the four optimization strategies with the hold-out test set. Performances were evaluated with the weights obtained at different epochs. 
We aim to evaluate if our method outputs better representations during training. In Figure~\ref{4methods_comparison}, we show performances (mean and standard deviation) on three different test sets for both datasets.
We also trained the encoder on the classification task in a fully supervised setting with 10\% and 100\% labeled data. For the fully supervised training, we used data augmentation composing the tested transformations randomly. Each transformation had a 0.5 probability of being sampled.
We performed linear evaluation on the frozen encoder with the hold-out test set and report the obtained AUC as horizontal lines in Fig.~\ref{4methods_comparison}.
Fig.~\ref{4methods_comparison} also reports linear evaluation results of the base simCLR optimization as in \cite{pmlr-v119-chen20j} where only one image is transformed by a random composition of the tested transformations. As with the fully supervised experiments, each transformation had a 0.5 probability of being sampled.

Fig.~\ref{4methods_comparison} shows that optimizing $\perturbator$ with supervision helps to have better representations for both datasets. It also shows that optimizing with only 10\% of labeled data allows us to reach the same quality of representation as the fully supervised training with 100\% of labels. 

To investigate the impact of the supervised loss function, we launched an experiment with the supervised contrastive loss introduced in~\cite{NEURIPS2020_d89a66c7} using only 10\% of labeled data. After convergence, we obtained a mean AUC of $0.52 \pm 0.12$  compared to $0.93 \pm0.01$ with our method.

On the Chest X-ray database, strong results were obtained in~\cite{Tang2020AutomatedAC} using a network pretrained on ImageNet. Optimizing $\perturbator$ with 10\% supervision on this ImageNet pretrained network has a smaller impact compared to random transformations ($0.96 \pm 0.001$ for both approaches). However, ImageNet pretrained networks can only be used with 2D slices whereas our strategy could be easily extended to 3D volumes.

\begin{figure}[h!]
\centerline{\includegraphics[width=0.8\linewidth]{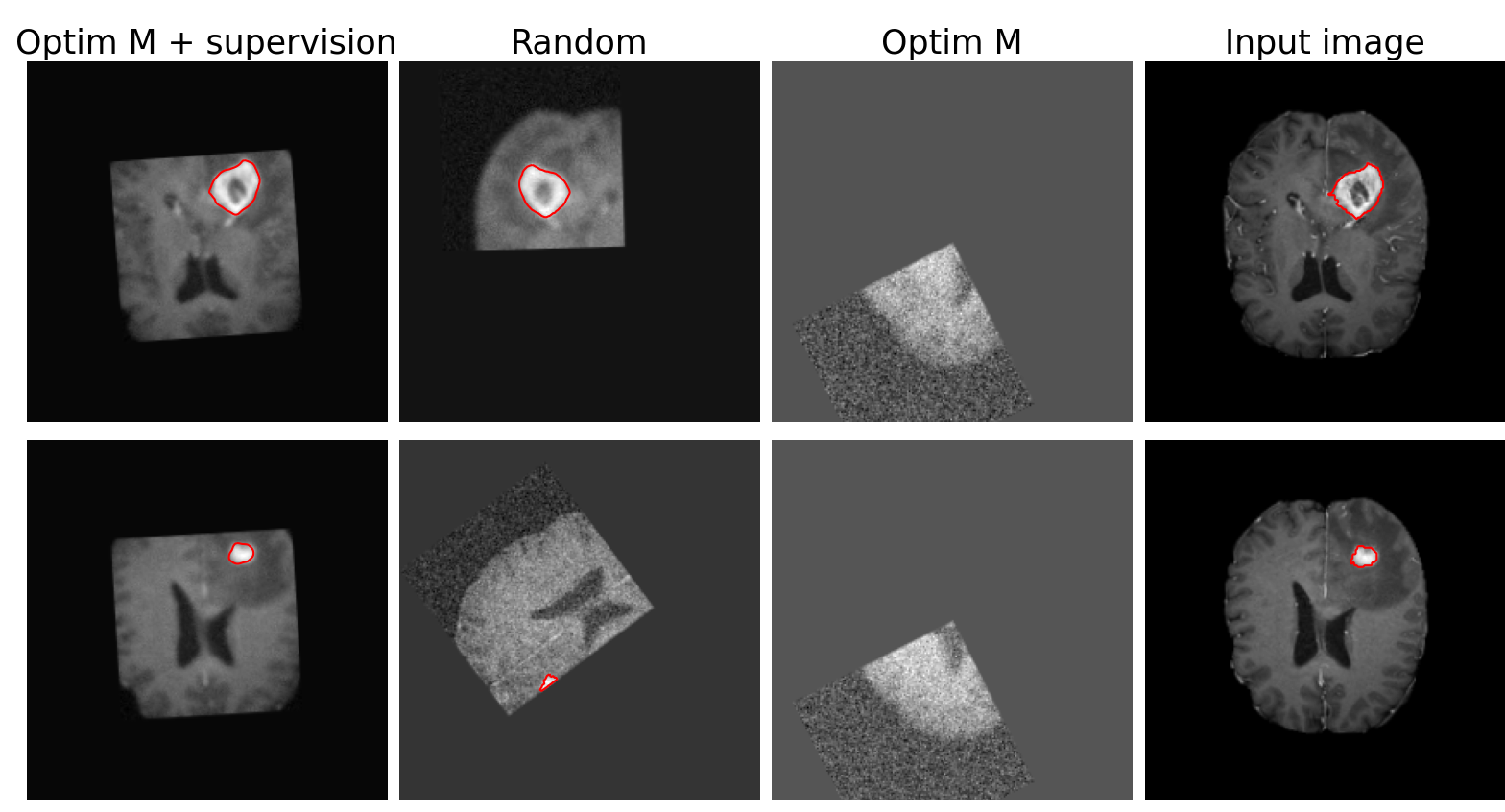}}
\caption{Two examples (row 1 and 2) of generated transformations in the BraTs dataset with different optimization strategies (red contour highlights the tumor).}
\label{perturbation_plot_Menc}
\end{figure}

\noindent \textbf{Relevance} When optimizing without supervision, the network $\perturbator$ needs to minimize the mutual information and it can therefore generate transformations that create images that are very different from the untransformed images but that do not contain relevant information for the downstream task, in particular for medical images. Without the supervision constraint, the optimal crop can be found, for instance, in a corner, leading to an image with a majority of zero values (i.e., entirely black), thus useless for the supervised task.
The supervision constraint helps $\perturbator$ to generate relevant images that keep pathological pixels (see some examples in Figure~\ref{perturbation_plot_Menc}).

\noindent \textbf{Runtime} 
The addition of the network $\perturbator$ increases the training computational time of around 20-25\% which is balanced by a performance gain.

\noindent \textbf{transformation composition order}
As in~\cite{pmlr-v119-chen20j}, the transformation order is fixed.
We launched an additional experiment with a different transformation order for both simCLR and our method. Linear evaluation results after convergence are respectively: 0.730$\pm$0.020 and  0.760$\pm$0.027 for simCLR and 0.926$\pm$0.020 and 0.923$\pm$0.021 for our method.
The transformation order has thus little impact on our results and, above all, our method substantially outperforms simCLR in both experiments.

\section{Conclusions and Perspectives}
    We proposed a method to optimize usual transformations employed in contrastive learning with very little supervision. 
    Extensive experiments on two datasets showed that our method finds more relevant transformations and obtains better latent representations, in terms of linear evaluation.
    Future works will try to optimize the transformations composition order. Furthermore, in a weakly-supervised setting, we could also investigate constraining latent space representations of non labeled data with pseudo-labels and nearest neighbor clustering.
%
%
%
\newpage
 \bibliographystyle{splncs04}
 \bibliography{ref.bib}

\end{document}